\documentclass{article} 
\usepackage{iclr2024_conference,times}

\usepackage{amsmath,amsfonts,bm}

\def\eqref#1{equation~\ref{#1}}

\def\1{\bm{1}}

\DeclareMathAlphabet{\mathsfit}{\encodingdefault}{\sfdefault}{m}{sl}
\SetMathAlphabet{\mathsfit}{bold}{\encodingdefault}{\sfdefault}{bx}{n}

\usepackage{hyperref}
\usepackage{url}
\usepackage{times}
\usepackage{latexsym}
\usepackage{float}
\usepackage{amsthm}
\usepackage{tcolorbox}
\usepackage{microtype}
\usepackage{amsmath, amssymb}
\usepackage{booktabs}
\usepackage{wrapfig}

\usepackage{graphicx}
\usepackage{multirow}

\usepackage{bm}
\usepackage{algorithm}
\usepackage{pifont}
\usepackage{subfigure}
\usepackage{xspace}

\title{Making Large Language Models Better Reasoners with Alignment}

\iclrfinalcopy

\author{Peiyi Wang$^1$ \quad Lei Li$^3$ \quad Liang Chen$^1$ \quad Feifan Song$^1$\\
\textbf{Binghuai Lin$^2$  \quad  Yunbo Cao$^2$ \quad Tianyu Liu$^2$ \quad Zhifang Sui$^1$}
\\ 
$^1$ National Key Laboratory for Multimedia Information Processing, Peking University\\
$^2$ Tencent Cloud AI \\
$^3$ The University of Hong Kong \\
 \texttt{\{wangpeiyi9979, nlp.lilei\}@gmail.com}  \\
 \texttt{leo.liang.chen@outlook.com; songff@stu.pku.edu.cn} \\
 \texttt{\{binghuailin, yunbocao, rogertyliu\}@tencent.com; \texttt{szf@pku.edu.cn}} 
}

\begin{document}

\maketitle

\begin{abstract}
Reasoning is a cognitive process of using evidence to reach a sound conclusion.
The reasoning capability is essential for large language models (LLMs) to serve as the brain of the artificial general intelligence agent.
Recent studies reveal that fine-tuning LLMs on data with the chain of thought (COT) reasoning process can significantly enhance their reasoning capabilities. 
However, we find that the fine-tuned LLMs suffer from an \textit{Assessment Misalignment} problem, i.e., they frequently assign higher scores to subpar COTs, leading to potential limitations in their reasoning abilities.
To address this problem, we introduce an \textit{Alignment Fine-Tuning (AFT)} paradigm, which involves three steps: 
1) fine-tuning LLMs with COT training data;
2) generating multiple COT responses for each question, and categorizing them into positive and negative ones based on whether they achieve the correct answer;
3) calibrating the scores of positive and negative responses given by LLMs with a novel constraint alignment loss.
Specifically, the constraint alignment loss has two objectives:
a) Alignment, which guarantees that positive scores surpass negative scores to encourage answers with high-quality COTs;
b) Constraint, which keeps the negative scores confined to a reasonable range to prevent the model degradation.
Beyond just the binary positive and negative feedback, the constraint alignment loss can be seamlessly adapted to the ranking situations when ranking feedback is accessible.
Furthermore, we also delve deeply into recent ranking-based alignment methods, such as DPO, RRHF, and PRO, and discover that the constraint, which has been overlooked by these approaches, is also crucial for their performance.
Extensive experiments on four reasoning benchmarks with both binary and ranking feedback demonstrate the effectiveness of AFT.
In addition, AFT also performs well in multi-task and out-of-distribution situations.

\end{abstract}

\section{Introduction}
Reasoning is a cognitive process that involves utilizing evidence to reach a well-founded conclusion \citep{qiao-etal-2023-reasoning,huang-chang-2023-towards}.
Recently, there has been a growing focus on enhancing the reasoning abilities of Large Language Models (LLMs) \citep{li-etal-2023-making}, particularly open-source LLMs \citep{rft,luo2023wizardmath,mukherjee2023orca}, because LLMs still lack reasoning skills \citep{wang2023large,wang2023far,zheng2023judging} that are essential for them to serve as the brain of artificial general intelligence agents \citep{wang2023voyager,react2023,song2023restgpt}.

\begin{figure}
    \centering
    \includegraphics[width=1\linewidth]{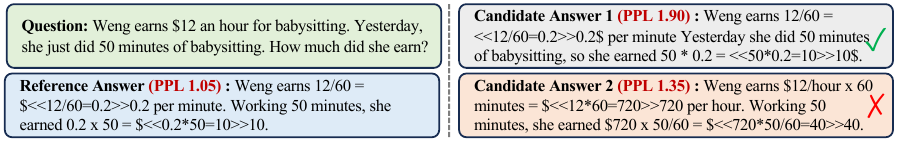}
    \vspace{-0.2in}
    \caption{Perplexity of different answers given by the vanilla fine-tuning (VFT) LLM, where LLM assigns a lower perplexity to the incorrect candidate answer compared to the correct candidate answer.}
    \vspace{-0.15in}
    \label{fig:intro}
\end{figure}

Recent works \citep{chung2022scaling,hsieh-etal-2023-distilling,mukherjee2023orca} find that training LLMs using data with a chain of thought (COT) reasoning process is a very effective method to improve the reasoning ability of LLMs.
These studies typically train LLMs using maximum likelihood estimation (MLE), and employ a next-token prediction objective.
However, MLE only assigns probability mass to the reference COT, which contradicts reasoning tasks where various reasoning paths can lead to the correct answer. 
In this paper, we find that previous vanilla fine-tuning (VFT) paradigm causes LLMs to suffer from an \textit{Assessment Misalignment} problem, i.e., LLMs struggle with accessing the quality of different COTs, ultimately limiting their reasoning capabilities.
Take Figure \ref{fig:intro} as an example, 
VFT-LLMs learn to generate the \textit{Reference Answer} for the given \textit{Question} by allocating probability mass to this \textit{Reference Answer} and treating all other answers as negative outcomes. As a result, they struggle to assess the quality of other answers and tend to assign lower perplexity (higher score) to \textit{incorrect Candidate Answer 1} compared to the \textit{correct Candidate Answers 2}.

This behavior of VFT-LLMs is not consistent with that of humans, as humans have the ability to access the quality of different COTs after learning to reason.
In addition, our pilot experiments (Section \ref{sec:pilot}) find that after the same VFT process, the LLMs with better reasoning performance can give a more reasonable assessment to different COTs.
Therefore, we hypothesize that we can improve the reasoning ability of LLMs by alleviating the assessment misalignment problem caused by VFT.

To address the assessment misalignment problem, in this paper, we propose an alignment fine-tuning (AFT) paradigm to improve LLM reasoning with three steps:
\textbf{1)} fine-tuning LLMs using COT training data;
\textbf{2)} generating multiple COT responses for each question using the fine-tuned LLMs, and categorizing them as positive and negative based on whether they deduce the correct answer;
\textbf{3)} calibrating the scores of positive and negative responses given by LLMs with a novel constraint alignment (CA) loss.
Specifically, the CA loss ensures that all positive scores (the scores of positive COTs) are larger than negative scores.
In addition, the negative scores are protected by a constraint term, which is proven to be very important in preventing model degradation.
Beyond just binary positive and negative feedback, the CA loss can be seamlessly adapted to ranking situations when ranking feedback is accessible.
Furthermore, we also delve deeply into recent ranking-based methods for alignment, such as DPO \citep{dpo}, PRO \citep{song2023preference} and RRHF \citep{yuan2023rrhf}, and find that the constraint, which has been overlooked by these approaches, is also crucial for their effectiveness.

In summary, our contributions are:

 \textbf{1)} We discover that LLMs fine-tuned by the vanilla fine-tuning (VFT) paradigm suffer from an Assessment Misalignment problem: they frequently assign lower scores to high-quality COTs compared to low-quality ones, which hinders their reasoning ability.

\textbf{2)} We present an Alignment Fine-Tuning (AFT) paradigm, which comprises three straightforward steps with a novel constraint alignment loss to address the identified problem.

\textbf{3)} We delve deeply into recent ranking-based methods for alignment and find that the constraint, which has been overlooked by these approaches, is also crucial for their performance.

\textbf{4)} Experiments on four reasoning benchmarks with both binary and ranking feedback demonstrate the effectiveness of AFT. AFT also performs well in multi-task and out-of-distribution situations.

\section{Related Works}

\subsection{Improve Reasoning of Large Language Models}
Reasoning is a cognitive process that involves utilizing evidence to reach a well-founded conclusion, which is a core ability of LLMs to serve as the brain of the artificial general intelligence agent.
Researchers have proposed a lot of methods to improve the reasoning ability of LLMs, which can be broadly divided into three groups:
1) \textit{pre-training}: The pre-training methods pre-train the LLMs on a vast of unsupervised datasets, such as the pile \citep{pile}, the stack \citep{stack}, and so on,  with a simple next token prediction objective. Researchers find that a larger model pre-trained on more data tends to have better reasoning ability  \citep{gpt4,palm,llama};
2) \textit{fine-tuning}: The fine-tuning methods can also enhance the reasoning ability of LLMs. Researchers have found that fine-tuning LLMs on the data with the reasoning chain-of-thought process can significantly improve the reasoning of LLMs \citep{mukherjee2023orca,chung2022scaling,li2023m};
3) \textit{prompting}: The prompting methods aims to improve the reasoning ability by carefully designed prompting strategy, such as chain-of-thought prompting \citep{cot}, self-consistency \citep{self-consistency} strategy, and so on. The prompting methods do not change the model parameters, which is very convenient and practical.
In this paper, we focus on the fine-tuning methods and find that traditional vanilla chain-of-thought fine-tuned LLMs suffer from an assessment misalignment problem, which hinders their reasoning ability.
To this end, we propose an alignment fine-tuning paradigm to address this problem to enhance the reasoning ability of LLMs.

\subsection{Alignment of Large Language Models}
AI alignment research focuses on directing AI systems toward human-intended goals, preferences, or ethical principles. There are two primary categories of AI alignment methods:
1) \textit{Reinforcement Learning from Human Feedback (RLHF)} \citep{instructionGPT}, which trains a reward model by utilizing human feedback, which subsequently acts as a reward function for optimizing an agent's policy through reinforcement learning (RL) techniques, such as Proximal Policy Optimization \citep{ppo}. RLHF is employed to align powerful LLMs, like ChatGPT and GPT-4. However, RL-based methods face limitations concerning training efficiency and complexity;
2) \textit{Supervised Fine-tuning with Ranking} \citep{brio,yuan2023rrhf,song2023preference,dpo}, which involves training LLMs using a supervised fine-tuning paradigm and incorporating a ranking loss to help LLMs align with human preferences.
Previous alignment research has mainly focused on improving the safety of LLMs, frequently neglecting the importance of alignment for reasoning. 
Furthermore, widely used ranking methods often neglect the constraint term when reducing scores of low-quality examples, which can potentially have a negative impact on model performance.
In this paper, we point out the effectiveness of alignment for reasoning and introduce a novel constraint alignment loss to make LLMs better reasoners with alignment.

\section{Pilot Experiments}
\label{sec:pilot}
In this section, we first briefly introduce the vanilla fine-tuning (VFT) paradigm, and then we demonstrate the assessment misalignment problem of VFT for reasoning.
\subsection{Vanilla Fine-tuning}
VFT finetunes LLMs on a dataset $\{(q_i, c_i, a_i)\}_{i=1}^N$ with $N$ examples.
Each example consists of a question $q_i$, a COT reasoning process $c_i$, and an answer $a_i$.
The LLMs are finetuned to generate the reference response $r_i=[c_i; a_i]$ based on $q_i$ with a MLE objective loss function:
\begin{equation}
\mathcal{L}_{VFT}=-\sum_{j=1}^{|r_i|} \log P(r_{i,j} \mid r_{i,<j}, q_i;\theta).
\label{eq:mle}
\end{equation}
where $\theta$ is the model parameter and $r_{i,j}$ is the $j$-th token of $r_i$.

\subsection{Assessment Misalignment of VFT for Reasoning}
Intuitively, the MLE objective seeks to exclusively allocate probability mass to the reference COT $c_i$ for question $q_i$, which does not correspond with the characteristics of reasoning tasks, where the correct COT is not limited to the reference one. This objective uniformly treats all other correct and incorrect COTs as negative examples. 
As a result, it will impede LLMs from learning to assess the quality of various COTs and degrade their reasoning ability.

To demonstrate this, we first fine-tune LLama-7B, LLama-13B, LLama2-7B, and LLama2-13B on the training data of GSM8k and ECQA with Equation \ref{eq:mle} (please refer to Section \ref{sec:param} for the detailed VFT settings). 
Then, for each question $q_i$ in the training data, we use VFT-LLMs to generate three positive COTs $\{c^{p1}_i, c_i^{p2}, c_i^{p3}\}$ that induce to the correct answer and three negative COTs  $\{c^{n1}_i, c_i^{n2}, c_i^{n3}\}$ that induce to the incorrect answer, respectively. 
Upon manually examining 50 examples, we observe that the quality of positive COTs is noticeably better than that of negative COTs.

We further compute the token-averaged log-likelihood score of each positive and negative COT $c$ using the fine-tuned LLMs as follows:
\begin{equation}
s_{\theta}^{c}=\frac{1}{|c|}{\sum_{j=1}^{|c|} \log P\left(c_j \mid c_{<j}, q;\theta\right)},
\label{eq:score}
\end{equation}
where $q$ is the corresponding question.
It is reasonable to expect that the fine-tuned LLMs will be able to assess the quality of different candidate COTs of previously encountered questions, i.e., assigning higher scores to the positive ones.
Therefore, we use an assessment accuracy $\mathbf{A_{Accuracy}}$ to assess the capability of fine-tuned LLMs in assigning appropriate scores to various COTs:
\begin{equation}
    \mathbf{A_{Accuracy}} = \frac{1}{9N}\sum_{i=1}^N\sum_{j=1}^3\sum_{k=1}^3 \mathbb{I} (s_{\theta}^{c_i^{pj}} > s_{\theta}^{c_i^{nk}})
\end{equation}

As shown in Table \ref{tab:pilot}, the assessment accuracy of the VFT-LLMs falls short of expectations, with an average $\mathbf{A_{Accuracy}}$ of merely around $70\%$ on GSM8K and  $62\%$ on ECQA, respectively.
Note that this is a two-class classification problem where a random baseline can achieve the $50.00\%$ accuracy.
These results show that the assessment ability of VFT-LLMs is far from expected, as they cannot accurately discern the quality of various COTs of previously learned questions.
This behavior of VFT-LLMs is not consistent with that of humans, as humans have the ability to access the quality of different COTs after learning to reason.
In addition, we also notice that LLMs with stronger reasoning abilities have better assessment accuracy.
Specifically, the task accuracy and the assessment accuracy exhibit a strong positive correlation, with Pearson Correlation Coefficients of 0.93 and 0.98 at GSM8K and ECQA, respectively. 
This observation inspires us to improve the reasoning ability of LLMs by aligning their scoring behaviors with the golden standard assessment.

\begin{table}[t]
    \centering
    \begin{tabular}{lcccc}
    \toprule
    \multirow{2}{*}{\textbf{\textsc{Models}}} & \multicolumn{2}{c}{\textbf{\textsc{Gsm8k (Pearson = 0.93)}}} & \multicolumn{2}{c}{\textbf{\textsc{Ecqa (Pearson = 0.98)}}} \\
     \cmidrule(r){2-3} \cmidrule(r){4-5} 
     & $\mathbf{T_{Accuracy}}$(\%) & $\mathbf{A_{Accuracy}}$(\%) & $\mathbf{T_{Accuracy}}$(\%) & $\mathbf{A_{Accuracy}}$(\%)  \\
    \midrule
       LLama-7B    & 36.48{$\pm$0.92} & 68.41{$\pm$0.32} & 70.40{$\pm$0.92} & 61.62{$\pm$0.01} \\
       LLama2-7B    & 40.71{$\pm$0.16} & 71.22{$\pm$0.12} & 72.34{$\pm$0.22} & 61.96{$\pm$0.02} \\
       LLama-13B  & 42.07{$\pm$0.15} & 72.25{$\pm$0.23} & 72.74{$\pm$0.43} & 61.89{$\pm$0.01}\\
       LLama2-13B  & 47.29{$\pm$1.24} & 73.06{$\pm$0.78} & 74.76{$\pm$0.56} & 62.29{$\pm$0.01} \\
    \bottomrule
    \end{tabular}
    \caption{The final task accuracy ($\mathbf{T_{Accuracy}}$) and the assessment accuracy ($\mathbf{A_{Accuracy}}$) of different vanilla fine-tuned models. $\mathbf{T_{Accuracy}}$ and $\mathbf{A_{Accuracy}}$ exhibit a strong positive correlation, with Pearson Correlation Coefficients of 0.93 and 0.98 at GSM8K and ECQA, respectively.}
    \vspace{-0.15in}
    \label{tab:pilot}
\end{table}

\section{Methodology}
We have demonstrated that the scoring behaviors of vanilla fine-tuned LLMs exhibit misalignment with the gold standard assessment.
In this section, we propose an alignment fine-tuning (AFT) paradigm to address this problem to enhance their reasoning ability.
Specifically, on top the VFT objective $\mathcal{L}_{VFT}$, AFT further introduce an alignment objective $\mathcal{L}_{A}^*$:
\begin{equation}
    \mathcal{L}_{AFT} = \mathcal{L}_{VFT} + \mathcal{L}_{A}^{*}.
\end{equation}
In the following part of this section, we will introduce the design process of $\mathcal{L}_{A}^{*}$. 
\subsection{Generate COTs for Training Data}

\label{sec:gen}
To implement AFT, we first need to generate multiple COTs for each question in the training set.
For each training example $(q, c, a)$, we first sample $k$ generation results $\{(c_i, a_i)\}_{i=1}^k$ from the VFT-LLMs based on the input question $q$.
Then, we divide these generation results into two groups, namely positive group $\mathbf{G}_P$ and negative group $\mathbf{G}_N$, based on the correctness of their answer.
Formally, a generation results $(c_i, a_i)$ belongs to $\mathbf{G_{P}}$ if $ a_i = a$, otherwise it is part of $\mathbf{G}_N$.
Generally, the quality of COTs in the positive group $\mathbf{G}_P$ is better than that of $\mathbf{G}_N$.

\subsection{Alignment}
As demonstrated by our pilot experiment, VFT-LLMs fail to give reasonable scores to COTs in $\mathbf{G_P}$ and $\mathbf{G_N}$.
To align the scoring behaviors of LLMs with the golden standard assessment, we need to design an objective to let the scores of all positive COTs in $\mathbf{G_P}$ larger than that of negative COTs in $\mathbf{G_N}$.
This objective bears resemblance to contrastive learning, which aims to ensure that the score of positive example is larger than those of all negative examples, utilizing an InfoNCE loss:
\begin{equation}
    \mathcal{L}_{InfoNCE} = -\log\left[\frac{\exp(s_{\theta}^{c_p})}{\exp(s_{\theta}^{c_p}) + \sum_{c_n \in \mathbf{G_N}} \exp(s_{\theta}^{c_n})}\right] = \log\left[1 + \sum_{c_n \in \mathbf{G_N}} \exp(s_{\theta}^{c_n} - s_{\theta}^{c_p})\right]
    \label{eq:infonce}
\end{equation}
Intuitively, minimizing Equation \ref{eq:infonce}  aims to make the positive score $s_{\theta}^{c_p}$ larger than all negative scores.
However, since there is more than one positive example in $\mathbf{G}_P$, inspired by \citep{su2022zlpr,wang-etal-2022-hpt}, we extend $\mathcal{L}_{InfoNCE}$ to accommodate multiple positive examples:
\begin{equation}
    \mathcal{L}_A= \log\left[1 + \sum_{c_p \in \mathbf{G_P}}\sum_{c_n \in \mathbf{G_N}}  \underbrace{\exp(s_{\theta}^{c_n} - s_{\theta}^{c_p})}_{\rm  alignment \ term} \right]
    \label{eq:c}
\end{equation}

where $s_{\theta}^c$ is the average log-likelyhood score of the COT $c$ calculated by Equation \ref{eq:score}.
Minimizing $\mathcal{L}_A$ encourages all positive scores to be larger than all negative scores.

\subsection{Constraint} 
Nevertheless, although the quality of negative COTs may not be as high as that of positive COTs, they still retain a respectable quality, as they are sampled from fine-tuned, powerful LLMs.
We find that reducing their scores by Equation \ref{eq:c} without setting any constraint will result in the degradation of the LLMs.
Therefore, we further design two constrained methods, Detached Constraint (DC), and Boundary Constraint (BC) to avoid such catastrophe.

\subsubsection{Detached Constraint}
To prevent model degradation, DC adds constraint to negative scores by detaching their gradient:
\begin{equation}
    \mathcal{L}^{DC}_A = \log\left[1 + \sum_{c_p \in \mathbf{G_P}}\sum_{c_n \in \mathbf{G_N}} \underbrace{\exp\left(\mathbf{D}(s_{\theta}^{c_n}) - s_{\theta}^{c_p}\right)}_{\rm \ detached \ alignment \ term}\right],
    \label{eq:dcc}
\end{equation}
where $\mathbf{D}(\cdot)$ denotes the detach operation, which means the gradient would not back-prop through the negative scores.
As a results, $\mathcal{L}^{DC}_A$ achieves the alignment by only increasing positive scores without explicitly decreasing negative ones.

\subsubsection{Boundary Constraint}
Besides DC, we also want to explore whether better results can be obtained by marginally decreasing negative scores. 
To this end, we propose BC that adds a constraint term to $\mathcal{L}_A$:
\begin{equation}
    \mathcal{L}_A^{BC} = \log\left\{1 + \sum_{c_p \in \mathbf{G_P}}\sum_{c_n \in \mathbf{G_N}} \left[\underbrace{\exp(s_{\theta}^{c_n} - s_{\theta}^{c_p})}_{\rm alignment \ term} +
 \underbrace{\exp(T - s_{\theta}^{c_n})}_{\rm constraint \ term}\right]\right\}
    \label{eq:BCA}
\end{equation}
Intuitively, the constraint term increases the score of the negative COT $s_{\theta}^{c_n}$, with the extent of improvement regulated by the value of $T$.
We aim for $T$ to achieve the effect of increasing $s_{\theta}^{c_n}$ when it is lower than a boundary $B$.
In this paper, we chose $B$ as the minimum positive COT score minus a hyper-parameter $\beta$, i.e., $B = s_{\theta}^{c_p*} - \beta$, where $ s_{\theta}^{c_p*}  = \min_{c_p \in \mathbf{G_P}}s_{\theta}^{c_p}$.
To achieve this, we analyze the gradient of Equation \ref{eq:BCA} with respect to the parameters $\theta$:
\begin{equation}
\begin{aligned}
      \nabla_{\theta}& \mathcal{L}_A^{BC} \propto - \sum_{c_p \in \mathbf{G_P}}\sum_{c_n \in \mathbf{G_N}} \left[ \exp(s_{\theta}^{c_n} - s_{\theta}^{c_p})(\nabla_{\theta}s_{\theta}^{c_p} -
     \nabla_{\theta}s_{\theta}^{c_n}) +  \exp(T - s_{\theta}^{c_n})\nabla_{\theta}s_{\theta}^{c_n} \right] \\
     =&-\sum_{c_p \in \mathbf{G_P}}\sum_{c_n \in \mathbf{G_N}} \left\{\underbrace{\exp(s_{\theta}^{c_n} - s_{\theta}^{c_p})\nabla_{\theta}s_{\theta}^{c_p}}_{\rm increase \ s_{\theta}^{c_p}} + \underbrace{ \left[\exp(T - s_{\theta}^{c_n}) -\exp(s_{\theta}^{c_n} - s_{\theta}^{c_p})\right] \nabla_{\theta}s_{\theta}^{c_n}}_{\rm change \ s_{\theta}^{c_n} \ based \ on \ the \ coefficient}\right\}
     \label{eq: gradient}
\end{aligned}
\end{equation}
Because the score $s_\theta^{c_*}$ increases along the gradient $\nabla_{\theta}s_{\theta}^{c_*}$,
based on $\nabla_{\theta} \mathcal{L}_A^{BC}$, for each pair $(c_p, c_n)$, $\mathcal{L}_A^{BC}$ consistently increases $s_\theta^{c_p}$ due to the positive coefficient $\exp(s_{\theta}^{c_n} - s_{\theta}^{c_p} ) > 0$.
Additionally, it elevates the negative score
$s_{\theta}^{c_n}$ when:
\begin{equation}
    \begin{split}
     \exp(s_{\theta}^{c_n} - s_{\theta}^{c_p})  < \exp(T - s_{\theta}^{c_n})
    \Rightarrow s_{\theta}^{c_n} < \frac{T+s_{\theta}^{c_p}}{2} = B
    \label{eq:increse_condition}
    \end{split}
\end{equation}
Otherwise, it tends to decrease or keep the score of $s_{\theta}^{c_n}$. Combing $B = s_{\theta}^{c_p*} - \beta$ and Equation \ref{eq:increse_condition}, we can achieve the value of $T=2s_{\theta}^{c_p*}-2\beta-s_{\theta}^{c_p}$. 

\subsection{Extending to Ranking alignment}
\label{sec: ranking}
The quality of the different COTs is not a simple binary relationship $\mathbf{G_P} \succ \mathbf{G_N}$, i.e., the quality of positive COTs is better than that of negative COTs.
In a more general situation, COTs in each group can also have quality differences, which means the quality of all generated COTs can be ranked as a sequence $c_1 \succeq c_2 \succeq \dots \succeq c_k$. 
If we can obtain such a quality ranking sequence, we can easily extend our binary-feedback boundary constraint alignment loss $\mathcal{L}_A^{BC}$ to a ranking-feedback boundary-constrained alignment loss as follows:
\begin{equation}
    {L}_A^{RBC} = \log\left\{1 + \sum_{c_i \succ c_j } \left[\underbrace{\exp(s_{\theta}^{c_j} - s_{\theta}^{c_i})}_{\rm alignment \ term} + \underbrace{\exp(2s^{c_j*}_{\theta} - 2\beta -  s^{c_i}_{\theta} - s_{\theta}^{c_j})}_{\rm constraint \ term}\right]\right\}
    \label{eq:rBCA}
\end{equation}
Where $s^{c_j*}_{\theta} = \min_{c_k \succ c_j} s_{\theta}^{c_k}$ is the minimal score of COTs that have the better quality than $c_j$.
Compared with $\mathcal{L}_A^{BC}$, ${L}_A^{RBC}$ can bring LLMs more detailed training signals of the COT assessment, which can further enhance their performance. 
We also try to extend $\mathcal{L}^{DC}_A$ to the ranking situation, and we find it slightly underperforms in comparison to ${L}_A^{RBC}$.
Please refer to Appendix \ref{sec:RDCC} for details.

\section{Experiments}
\subsection{Experimental Setups}
\paragraph{Datasets}
We conduct our experiments on three widely used reasoning datasets with human-annotated chain-of-thoughts, including math reasoning tasks \textbf{GSM8K} \citep{gsm8k}, \textbf{AQUA-RAT} \citep{aqua}, commonsense reasoning task \textbf{ECQA} \citep{ecqa}. 
Furthermore, we create \textbf{GSM8K-RANK} to evaluate the effectiveness of our AFT in the ranking situation.
Please refer to Appendix \ref{sec:datasets} for more details of these datasets.

\paragraph{Parameter Setting}
\label{sec:param}
We conduct experiments on four large language models, LLama(2)-7B and LLama(2)-13B.
We do not conduct experiments on larger models due to resource limitations.
We sample $k=6$ COTs from VFT-LLMs with a sampling temperature of 1. 
Our detached constraint alignment loss does not introduce any hyper-parameters, and we search the boundary constraint hyper-parameter $\beta$ based on the validation set.
For more training details, please refer to Appendix \ref{sec: param_setting}.

\paragraph{Baselines}
We compare our AFT with the following baselines: 
1) \textbf{VFT}: the vanilla fine-tuning (VFT) method that simply trains LLMs with the reference COT using the MLE loss, which is the most widely used training strategy; 
2) \textbf{RFT}: Rejective sampling fine-tuning (RFT) \citep{rft} selects the COTs with the correct answer, adds these COTs to the origin training data, and uses the new augmented training data to train LLMs, which is proven to be a very strong baseline;
3) \textbf{RRHF}: Rank Responses to align Human Feedback (RRHF) \citep{yuan2023rrhf}, which takes candidate ranking into account and distinguishes different candidates through a pair-wise ranking loss;
4) \textbf{PRO}: Preference Ranking Optimization (PRO) \citep{song2023preference}, which takes candidate ranking into account and distinguishes different candidates through a ranking loss with a dynamic temperature.

\paragraph{Metrics} We use the accuracy to measure the model performance. Specifically, we conduct $3$ runs with $3$ different seeds and report the average results with the standard deviation.

\subsection{Results with Binary Feedback}

\begin{table*}[t]
    \centering
    \scalebox{0.95}{
    \begin{tabular}{clcccl}
    \toprule
    \textbf{\textsc{Models}} & \textbf{\textsc{Methods}}  & \textbf{\textsc{Gsm8k}} & \textbf{\textsc{Aqua}} & \textbf{\textsc{Ecqa}}   &  \textbf{\textsc{Average} ($\Delta$)} \\
    \midrule
     \multirow{4}{*}{\rotatebox{0}{ {\textsc{LLama-7B}} }} &  VFT  & 36.48{$\pm$0.92} & 31.19{$\pm$0.28} & 70.40{$\pm$1.07} &  46.02 ( \ \ \ -- \ \ \ ) \\
     & RFT & 39.75{$\pm$1.03} & 32.81{$\pm$1.48} & \textbf{72.23{$\pm$0.11}} & 48.28 ($\uparrow$ 2.26)\ \\
     \cmidrule(r){2-6}
     & AFT ($\mathcal{L}^{DC}_A$) & \textbf{40.43{$\pm$1.04}} & 33.01{$\pm$0.95} & \textbf{72.23{$\pm$0.43}} & \textbf{48.55 ($\uparrow$ 2.53)}\ \\
     & AFT ($\mathcal{L}_A^{BC}$)  & {40.26{$\pm$0.36}} & \textbf{33.20{$\pm$1.24}} &  {72.15{$\pm$0.57}} & {48.53} ($\uparrow$ 2.51)\\

    \midrule
     \multirow{4}{*}{\rotatebox{0}{ {\textsc{LLama2-7B}} }} &  VFT  & 40.71{$\pm$0.16} & 31.49{$\pm$1.96} &  {72.34{$\pm$0.22}} & 48.18 ( \ \ \ -- \ \ \ ) \\
     & RFT & {43.65{$\pm$0.13}} & {33.25{$\pm$1.23}}   & \textbf{73.86{$\pm$0.38}} & 50.25 ($\uparrow$ 2.07) \\
     \cmidrule(r){2-6}
     & AFT ($\mathcal{L}^{DC}_A$)   & \textbf{44.25{$\pm$0.43}} & \textbf{33.49{$\pm$0.63}} & 73.71{$\pm$0.65} & \textbf{50.75 ($\uparrow$ 2.57)} \\
     & AFT ($\mathcal{L}_A^{BC}$)  & 44.16{$\pm$0.81} & 32.89{$\pm$0.98} & {73.23{$\pm$0.82}} &  50.09 ($\uparrow$ 1.91)\\
    \midrule
       \multirow{4}{*}{\rotatebox{0}{ {\textsc{LLama-13B}} }} & VFT  & 42.07{$\pm$0.15}   & 33.91{$\pm$0.60} & 72.74{$\pm$0.43} & 49.57 ( \ \ \ -- \ \ \ )\\
       & RFT & 46.13{$\pm$1.41} & 34.29{$\pm$1.28}& \textbf{75.03{$\pm$0.35}} & {51.80 ($\uparrow$ 2.23)}  \\
       \cmidrule(r){2-6}
      & AFT ($\mathcal{L}^{DC}_A$)  & {46.31{$\pm$1.52}} & 34.49{$\pm$1.21} & 74.32{$\pm$0.09} & 51.70  ($\uparrow$ 2.13)\\
     & AFT ($\mathcal{L}_A^{BC}$)  & \textbf{46.46{$\pm$0.28}} & \textbf{34.79{$\pm$0.37}} & {74.53{$\pm$0.68}} &  \textbf{51.93 ($\uparrow$ 2.36)} \\

      \midrule
     \multirow{4}{*}{\rotatebox{0}{ {\textsc{LLama2-13B}} }} &  VFT  & 47.29{$\pm$1.24} & 34.68{$\pm$1.36}  & 74.76{$\pm$0.56} &  52.24  ( \ \ \ -- \ \ \ )  \\
     & RFT & 50.12{$\pm$1.57} & 34.95{$\pm$0.88} & {76.21{$\pm$0.80}}  &  53.75 ($\uparrow$ 1.51)\\
     \cmidrule(r){2-6}
      & AFT ($\mathcal{L}^{DC}_A$)  & 50.67{$\pm$1.16} & \textbf{35.78{$\pm$0.45}} & {76.42{$\pm$0.82}} & {54.29} ($\uparrow$ 2.05) \\
      & AFT ($\mathcal{L}_A^{BC}$)  &  \textbf{51.03{$\pm$0.54}} &  {35.49{$\pm$1.19}} & \textbf{76.57{$\pm$0.83}} & \textbf{54.36 ($\uparrow$ 2.12)} \\

    \bottomrule
    
    \end{tabular}}
    \caption{The accuracy of different methods on three reasoning datasets. $\Delta$ denotes the improvement compared to VFT. AFT significantly outperforms  VFT, and is slightly better than RFT \citep{rft}. Note that RFT is a concurrent work to ours.}
    \label{tab:main}
    \vspace{-0.1in}
\end{table*}

Table \ref{tab:main} displays the results of different fine-tuning methods on three reasoning datasets. 
As is shown:
\textbf{1)}: AFT significantly outperforms VFT on all three datasets, improving the average accuracy by $1.91\% \sim 2.57\%$ for all models,  showing the effectiveness of AFT;
\textbf{2)}: Our concurrent work RFT also expresses notable improvement compared with VFT.
However, the original RFT paper only treats RFT as a simple data augmentation method without explaining the reasons behind its notable improvement.
Our alignment perspective can provide an explanation for the effectiveness of RFT, i.e., RFT can alternatively be regarded as an alignment strategy that bolsters the scores of numerous positive COTs and thus can alleviate the assessment misalignment problem of VFT;
\textbf{3)} Our proposed two constraint alignment strategies slightly outperform RFT with the binary feedback.
In addition, our AFT can be also easily extended to utilize the ranking feedback that RFT can not well utilize.
These results demonstrate the importance of revealing the assessment misalignment problem of VFT and the effectiveness of our AFT approach.

\subsection{Results with Ranking Feedback}
\begin{table}[t]
    \centering
    \small
        \begin{tabular}{lccccc}
        \toprule
              \textsc{\textbf{Methods}} & \textsc{\textbf{LLama-7B}} & \textsc{\textbf{LLama-13B}} & \textsc{\textbf{LLama2-7B}} & \textsc{\textbf{LLama2-13B}} & \textsc{\textbf{Average}}  ($\Delta$)  \\
              \midrule
              VFT  & 20.82{$\pm$0.71} & 24.12{$\pm$0.42} & 24.08{$\pm$0.22} & 30.28{$\pm$1.46} & 24.83 ( \ \ \ -- \ \ \ ) \\
              RFT  & 25.09{$\pm$1.18} & 28.21{$\pm$0.86} & 28.25{$\pm$0.78} & 34.53{$\pm$0.51} & 29.02 ($\uparrow$ 4.19)\\
              RRHF & 7.51{$\pm$0.56} & 9.92{$\pm$0.82} & 9.21{$\pm$0.25} & 13.35{$\pm$1.26}  & 10.00 ($\downarrow$ 14.8)\\
              PRO &  18.73{$\pm$0.31} & 20.34{$\pm$1.51} & 21.40{$\pm$0.92} & 23.55{$\pm$0.98} & 21.00 ($\downarrow$ 3.82)  \\
              \midrule
              AFT (${L}_A^{RBC}$) & \textbf{26.08{$\pm$1.05}} & \textbf{28.97{$\pm$0.35}} & \textbf{29.05{$\pm$0.75}} & \textbf{35.48{$\pm$1.35}} & \textbf{29.90 ($\uparrow$ 5.07)}  \\  
        \bottomrule
        \end{tabular}
    \caption{Test accuracy of different methods on GSM8K trained with GSM8K-RANK.}
    \label{tab:rank}
\end{table}
As described in Section \ref{sec: ranking}, our AFT can also be easily adapted to the ranking situation where we can obtain the quality ranking sequence of generated COTs.
Table \ref{tab:rank} illustrates the results of different methods in the GSM8k-RANK. 
As is shown:
\textbf{1)} Our AFT surpasses all other methods, demonstrating its effectiveness with ranking feedback. For instance, AFT exceeds the strongest baseline RFT by 0.88\% in average accuracy.
This superiority can be attributed to AFT's ability to help LLMs recognize quality differences among any given pair in a ranking context, while RFT only focuses exclusively on optimizing the probability of the highest-quality examples;
\textbf{2)} Prior methods utilizing ranking loss have a substantial negative impact on model performance.
For example, integrating RRHF loss into VFT leads to a 14.8\% reduction in accuracy. 
In fact, the performance reduction is also observed in their own paper \citep{song2023preference}, which demonstrates that ranking loss often enhances the reward of LLMs, yet results in lower BLEU scores.
However, they do not identify the cause, and in this paper, we find that a potential reason for the performance decline is the absence of a constraint in their loss, which we will discuss in Section \ref{sec:other_rank}.

\section{Analysis}

\subsection{Delve Deeply into Recent Ranking Losses for Alignment}

\begin{table}[t]
    \centering
    \small
    \begin{tabular}{lcccccc}
    \toprule
       \multirow{2}{*}{\rotatebox{0}{ {\textsc{\textbf{Methods}}} }}  & \multicolumn{3}{c}{\textsc{\textbf{without constraint}}}  & \multicolumn{3}{c}{\textsc{\textbf{with constraint  (Ours)}}} \\
       \cmidrule(r){2-4} \cmidrule(r){5-7}
       & $\mathbf{T_{Accuracy}}$  & $\mathbf{A_{Accuracy}}$  & $\mathbf{PPL}$ ($\downarrow$) & $\mathbf{T_{Accuracy}}$  & $\mathbf{A_{Accuracy}}$ & $\mathbf{PPL}$ ($\downarrow$) \\
    \midrule
     VFT  & \textbf{20.82{$\pm$0.71}} & 68.72{$\pm$1.48}          & \textbf{1.60{$\pm$0.01}} &            20.82{$\pm$0.71} & 68.72{$\pm$1.48} & 1.60{$\pm$0.01} \\
    RRHF     & 7.51{$\pm$0.56} & 87.44{$\pm$1.28}        & 1.80{$\pm$0.01} &          25.53{$\pm$0.27} & 79.89{$\pm$0.60} & \textbf{1.35{$\pm$0.01}} \\
    PRO   & 18.73{$\pm$0.31} &  86.58{$\pm$1.09}         & 2.34{$\pm$0.02} &          25.82{$\pm$0.48}  & 80.34{$\pm$0.97} & 1.45{$\pm$0.01} \\
    \midrule
    AFT (${L}_A^{RBC}$)     & 7.03{$\pm$0.98} & \textbf{88.89{$\pm$0.78}} & {7.81$\pm$0.03} & \textbf{26.08{$\pm$1.05}} & \textbf{81.36{$\pm$0.78}} & 1.37{$\pm$0.01} \\
    \bottomrule
    \end{tabular}
    \caption{Task accuracy ($\mathbf{T_{Accuracy}}$) and assessment accuracy ($\mathbf{A_{Accuracy}}$) on GSM8K for LLama-7B, which is fine-tuned by different methods (with or without constraint) on GSM8K-Rank. $\mathbf{PPL}$ ($\downarrow$, lower is better) denotes the average perplexity of all positive COTs.}
    \label{tab:constraint}
    \vspace{-0.1in}
\end{table}

\label{sec:other_rank}
Our experiments on GSM8K-RANK show that adding ranking loss will harm the model performance. 
We think the reason is that previous alignment ranking losses will unreasonably decrease the score of non-optimal COTs (Please refer to Appendix \ref{sec:analysis_rank} for our detailed analysis).
To empirically validate this hypothesis, we add a detached constraint to these two ranking losses similar to $\mathcal{L}_A^{RDC1}$ (Equation \ref{eq:rRDC1}). 
Consequently, these ranking losses will only make the scores of higher-quality COTs larger than those of lower-quality ones, without explicitly decreasing the scores of COTs with lower quality. 
Table \ref{tab:constraint} illustrates the final accuracy $\mathbf{T_{Accucary}}$ of different methods in the testing set, the assessment accuracy
$\mathbf{A_{Accuracy}}$ and average perplexity of positive COTs $\mathbf{PPL}$ in the training set\footnote{For each question in the training set, we sample new COTs (three positive and three negative COTs, respectively) that are different from training COTs for evaluation.}.
As is shown:
\textbf{1)} Without the constraint strategy, all three ranking losses harm the model performance, leading to higher perplexity  and lower final task accuracy  compared to VFT;
\textbf{2)} We observe that the task accuracy of PRO does not decline as significantly as RRHF and AFT. We think this is because PRO employs a dynamic temperature that reduces the negative score in a more reasonable manner (Please refer to Appendix \ref{sec:pro} for details);
\textbf{3)} By adding the constraint, all ranking losses can not only improve two accuracies but also decrease the perplexity.
These results show the importance of constraint for other ranking losses for alignment.
Furthermore, we also conduct a case study in Appendix \ref{sec:case} to intuitively show the model degradation without constraint.

\subsection{Effectiveness of the Number of Candidate COTs} 
\label{sec:num_k}
\begin{figure}[t]
\centering
\subfigure{
\includegraphics[width=0.31\textwidth]{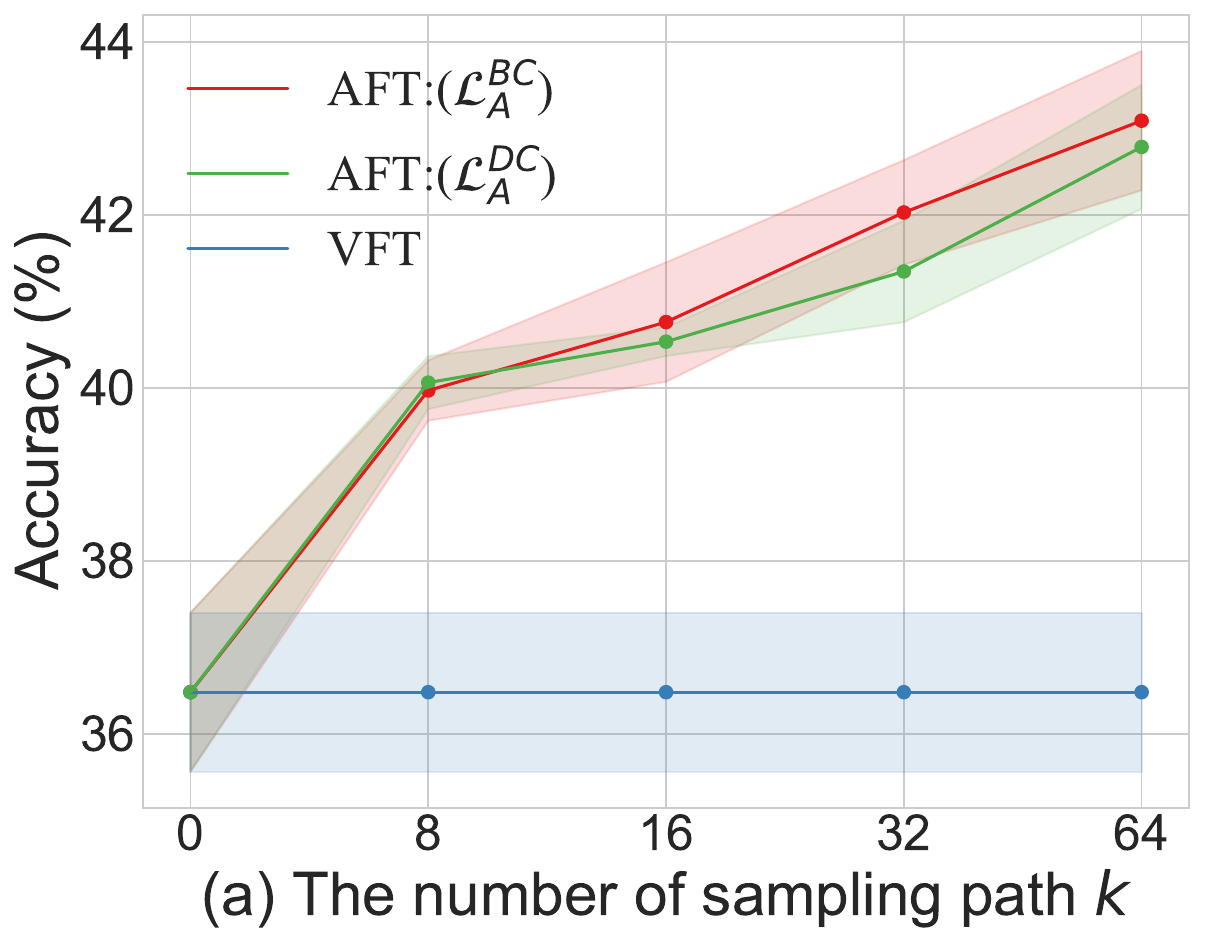}
}
\subfigure{
\includegraphics[width=0.31\textwidth]{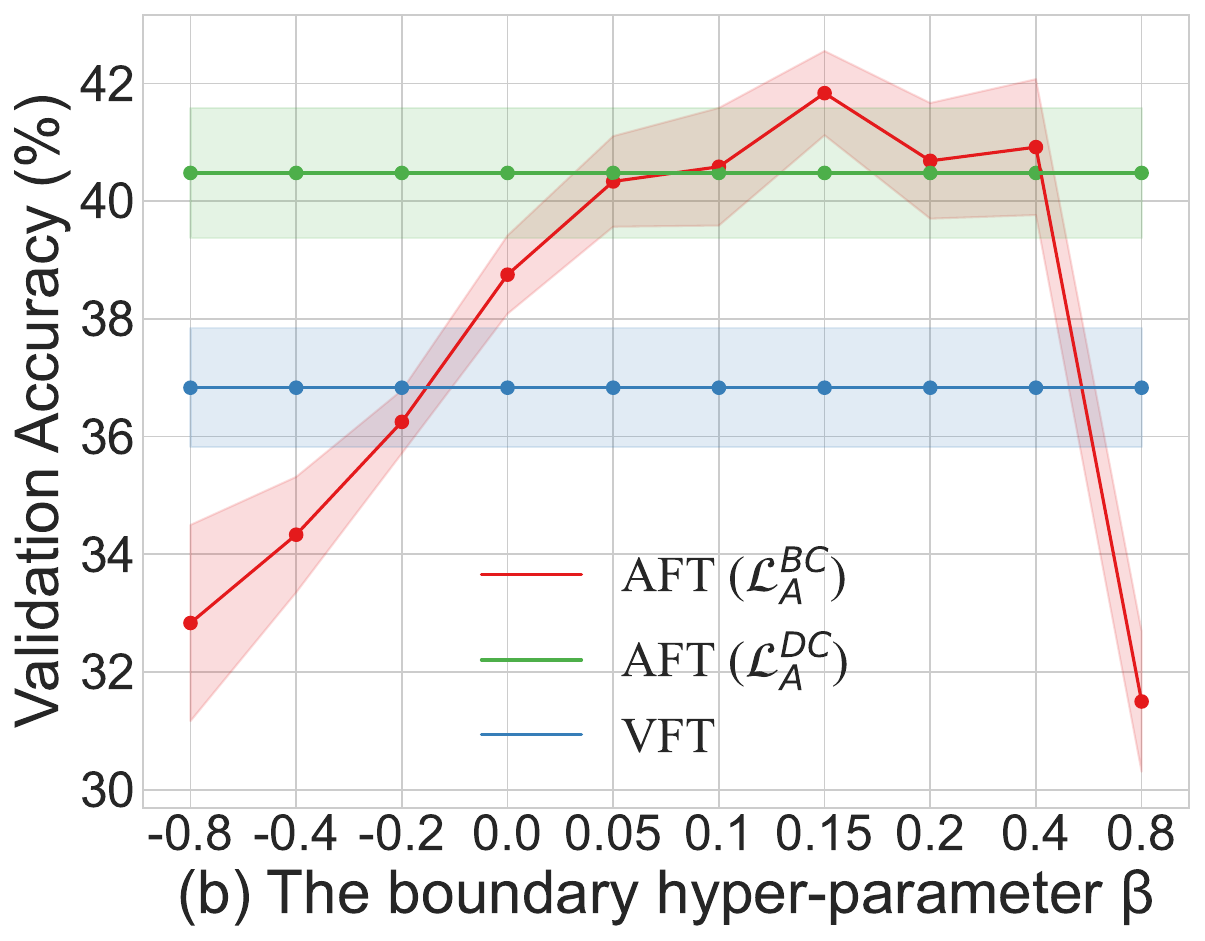}
}
\subfigure{
\includegraphics[width=0.31\textwidth]{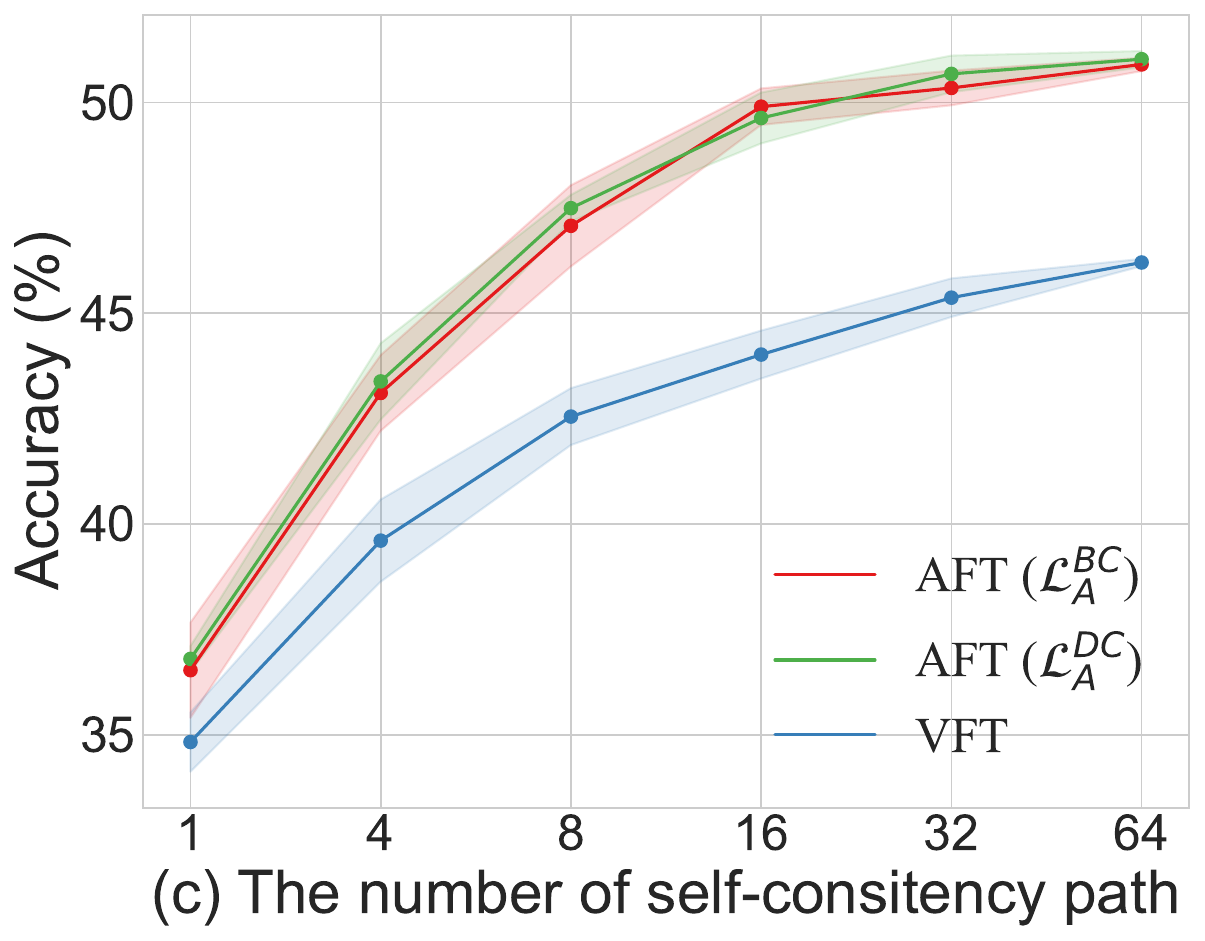}
}
\caption{Variation of accuracy with (a): different number of sampling COTs for training; (b) different boundary constraint hyper-parameter; (c) different number of voting paths of self-consistency.}
\label{fig:hyp}
\end{figure}

As described in Section \ref{sec:gen}, AFT samples $k$ candidate generation results to align LLMs.
In this section, we explore the influence of $k$.
We sampled 0, 8, 16, 32, and 64 results from the VFT-LLama-7B, and then de-duplicated these sampling results.
Then, we train LLama-7B on the de-duplicated datasets.
As shown in Figure \ref{fig:hyp}(a), we can see that AFT can consistently improve the model performance with $k$ improvement, which is a promising result.
We think the reason is that with large $k$, the AFT will have more data to help the LLM perceive the quality of different COT paths, which enhances the final performance.
This growing accuracy shows the effectiveness and the potential of AFT.

\begin{table}[t]
    \centering
    \begin{tabular}{lccccc}
    \toprule
        \textbf{\textsc{Methods}} & \textbf{\textsc{Gsm8k}}   & \textbf{\textsc{Aqua}} & \textbf{\textsc{Ecqa}} & \textbf{\textsc{Mmlu}} & \textbf{\textsc{Average}} ($\Delta$) \\
        \midrule
         \multirow{1}{*}{\rotatebox{0}{ {VFT} }} & 35.72{$\pm$0.95} & 32.95{$\pm$0.98}  & 69.25{$\pm$0.74} & 37.52{$\pm$1.03} & 43.86  ( \ \ \ -- \ \ \ ) \\
        \midrule
        \multirow{1}{*}{\rotatebox{0}{ {AFT ($\mathcal{L}^{DC}_A$) }}} & 40.24{$\pm$0.63} &	33.72{$\pm$0.92} &	71.38{$\pm$0.64} & 39.25{$\pm$0.35} & 46.15 ($\uparrow$ 2.29) \\
        \multirow{1}{*}{\rotatebox{0}{ {AFT ($\mathcal{L}^{BC}_A$) }}} & 40.00{$\pm$0.69} &	33.45{$\pm$0.56} & 71.48{$\pm$0.89} & 38.89{$\pm$0.70}  & 45.96 ($\uparrow$ 2.10)  \\
        
    \bottomrule
    \end{tabular}
    \caption{Comparison of VFT- and AFT-LLama-7B with training data ``GSM8K+AQUA+ECQA'' on three in-domain benchmarks and an out-of-domain benchmark MMLU.}
    \label{tab:multi}
    \vspace{-0.15in}
\end{table}

\subsection{Ablation on the Boundary Value}
The boundary constraint term of AFT requires a hyper-parameter $\beta$ to regulate the boundary. In this section, we conduct an ablation study to demonstrate the impact of varying $\beta$  values. As depicted in Figure \ref{fig:hyp}(b), the performance initially increases and subsequently decreases as $\beta$ ranges from -0.8 to 0.8. These findings align with expectations, as a small $\beta$ cannot effectively widen the score gap between high-quality and low-quality COTs, while an overly large $\beta$ may result in excessively low scores for non-optimal COTs, thereby compromising the model's generative abilities. 
In conclusion, the results emphasize the importance of the boundary constraint term and indicate that the value of $\beta$ can significantly affect model performance. 
Therefore, it is essential to carefully adjust this value when using our boundary constraint alignment loss.

\subsection{Effectiveness of AFT with Self-Consistency}

Self-consistency is a highly effective strategy for improving LLM's reasoning performance.  This method involves sampling multiple COTs and utilizing a voting process to determine the final answer during inference. AFT samples COTs for training to develop better LLMs. Both methods utilize COTs to enhance the model's reasoning ability.
In this section, we explore the combination of AFT and Self-Consistency. As illustrated in Figure \ref{fig:hyp}(c), as the number of paths increases, the improvement of AFT is more significant than VFT, demonstrating that AFT effectively enhances self-consistency. We believe the reason is that AFT helps models learn to assess the quality of different COTs by encouraging larger scores for high-quality COTs compared to low-quality ones. This means that high-quality COTs are more likely to be sampled, and thus, AFT can enhance self-consistency.

\subsection{Effectiveness of AFT on the Multi-task and Out-of-domain Situations} 

To further demonstrate the effectiveness and versatility of AFT, we investigate its performance in multi-task scenarios. We combine the training sets of three datasets and use both AFT and VFT to train the LLama-7B model. As depicted in Table \ref{tab:multi}, AFT is able to simultaneously enhance the performance of all corresponding test sets.
Additionally, we evaluate both AFT and VFT on the MMLU (zero-shot), an out-of-distribution benchmark, and
AFT also outperforms VFT.
These results indicate that AFT not only improves the performance of in-distribution tasks but also enhances the model's transfer ability, leading to significantly better out-of-distribution performance.

\section{Conclusion}
In this paper, we find that the vanilla fine-tuned (VFT) LLMs with chain-of-thought (COT) reasoning process suffer from an assessment misalignment problem, i.e, they fail to access the quality of different COTs of the learned questions, which hinders the reasoning ability of LLMs.
To this end, we propose an alignment fine-tuning (AFT) paradigm.
Our AFT consists of a novel constraint alignment loss that can align the model assessment behaviors without harming the model performance. 
Furthermore, we also delve deeply into recent widely used ranking losses for alignment and find that the constraint, which has been overlooked by these approaches, is also crucial for their performance.
Extensive experiments on four reasoning benchmarks demonstrate the effectiveness of AFT. 
In addition, AFT also performs well in multi-task and out-of-distribution situations.

\section{Limitations}
Our paper has some limitations, which should be discussed in future works:
\textbf{1)} Due to the resource limit, we do not scale the AFT to larger LLMs such as 65B and 70B LLama models. However, we believe that larger models still suffer from the assessment misalignment problem of VFT, and thus AFT can improve the performance of these larger models;
\textbf{2)} Our boundary constraint alignment loss incorporates a hyper-parameter $\beta$ that regulates the constraint strength, significantly impacting the model's performance. Finding the optimal hyper-parameter requires constructing a validation set and a certain search overhead. 
Although our detached alignment loss can mitigate the assessment misalignment problem without requiring any hyper-parameters, it sometimes falls short in comparison to the boundary constraint alignment loss, especially in ranking situations. Therefore, how to design a dynamic boundary constraint without introducing the hyper-parameter is a meaningful question, which leaves for further work.

\bibliography{AFT}
\bibliographystyle{iclr2024_conference}

\clearpage
\appendix

\section{Datasets}

\begin{table}[t]

\begin{tcolorbox}
I want you to act as a grade school math teacher, and evaluate the quality of the answer provided by an AI assistant to the math Question displayed below.

You will be given a reference answer and the assistant's answer, and Your evaluation should consider the correctness of the assistant's answer.

Begin your evaluation by comparing the assistant's answer with the reference answer step-by-step. Identify and correct any mistakes. 

The answer is scored out of 10 points, with one point deducted for each wrong step. Be as objective as possible.

Your need first provide your Evaluation Evidence and then rate the response on a scale of 1 to 10.

[Question]: 

\textcolor{red}{\{question\}}

[The Start of Reference Answer]

 \textcolor{red}{\{reference\}}
 
[The End of Reference Answer]

[The Start of Assistant's Answer]

 \textcolor{red}{\{answer\}}
 
[The End of Assistant's Answer]

You MUST output with two lines:

Evaluation Evidence: $<$Explanation$>$

Rating: $<$ONLY a single digit$>$

\end{tcolorbox}
\caption{The evaluation template that prompts ChatGPT to score each candidate COT.}
\label{tab:template}
\end{table}

\label{sec:datasets}
We conduct our experiments on three widely used reasoning datasets with human-annotated chain-of-thoughts, including math reasoning tasks GSM8K \citep{gsm8k}, AQUA-RAT \citep{aqua}, commonsense reasoning task ECQA \citep{ecqa}:

\noindent\textbf{GSM8K} GSM8K is a widely used mathematical reasoning dataset, which comprises 8.5K varied math word problems for grade school, developed by human authors. It is partitioned into 7.5K training problems and 1K testing problems. We sample 400 problems from the testing set to form the validation set, and thus we have $7,473$, $400$, and $919$ examples in training, validation, and testing sets, respectively.

\noindent\textbf{AQUA-RAT} AQUA-RAT comprises approximately $100,000$ algebra-based word problems, each accompanied by a natural language rationale. Each example in the dataset consists of four components: 1) question, which statement is written in natural language, 2) options, a set of five potential answers with one being correct, 3) rationale, a natural language explanation of the problem's solution, and 4) correct, the right answer choice.
For efficiency, we randomly sample $5,000$, $400$, and $1,254$ examples as the training, validation, and test set, respectively.

\noindent\textbf{ECQA} ECQA is derived from CommonsenseQA (CQA) \citep{csqa} by generating a free-flow explanation for each QA pair in CQA. 
CQA is a comprehensive dataset for commonsense reasoning, containing QA pairs with five choices and a single correct answer. ECQA comprises 11K QA pairs in total and has $7,598$, $1,090$, and $2,194$ examples in the training, validation, and test sets, respectively.

\noindent\textbf{GSM8K-RANK}  To evaluate the effectiveness of our AFT in the ranking situation, we randomly select 1,000 examples from GSM8K's training set and generate 8 candidate COTs for each question. 
We then prompt ChatGPT to rate these candidates by providing the question, reference answer, and the COT to be assessed and thus we can achieve a quality ranking sequence for different generated COTs. 
We randomly sampled 20 examples and found that ChatGPT's scoring results align well with human assessment.
ChatGPT is instructed to assign a score between 1 and 10, indicating the quality of each COT. To ensure the reliability of the ratings, following \citep{wang2023large}, we require ChatGPT to present evaluation evidence before assigning a score, and simple 3 scores for each example.
We take the average score as the final score for each COT. 

\section{Parameter Setting}
\label{sec: param_setting}
\begin{table}[t]
    \centering
    \begin{tabular}{lcccc}
    \toprule
       Models  & GSM8K & AQUA & ECQA & GSM8K-RANK \\
    \midrule
       LLama-7B  & 0.15 & 0.15 & 0.15 & 0.05\\
       LLama2-7B & 0.15 & 0.40 & 0.35 & 0.15 \\
       LLama-13B & 0.15 & 0.15 & 0.15 & 0.15 \\
       LLama2-13B & 0.15 & 0.15 & 0.20 & 0.15\\
    \bottomrule
    \end{tabular}
    \caption{The value of hyper-parameter $\beta$ for boundary constraint alignment.}
    \label{tab:beta}
\end{table}
We conduct experiments on four large language models, LLama-7B, LLama-13B, LLama2-7B, and LLama2-13B.
We do not conduct experiments on larger models due to resource limitations.
We sample $k=6$ COTs from VFT-LLMs with a sampling temperature of 1. 
Our detached constraint alignment loss does not introduce any hyper-parameters, and we search the hyper-parameter of boundary constraint loss within the range $(0, 0.05, 0.1, 0.15, 0.2, 0.25, 0.3, 0.35, 0.4, 0.45, 0.5)$ on the validation set.
The value of $\beta$ of different models and datasets is provided in Table \ref{tab:beta}.
On GSM8K, AQUA, and ECQA, the models are trained for 3, 3, and 1 epochs, respectively.
The learning rate is set to 2e-5, featuring linear decay and a linear warmup for 3\% of the total training steps. 
7B and 13B models are trained on 8 and 32 V100 GPUs with 32GB memory, respectively. 
We employ a maximum sequence length of 512 and utilize the DeepSpeed library and ZeRO optimizer during training.

\section{detached constraint Ranking Loss}
\label{sec:RDCC}
Given a ranking sequence $c_1 \succeq c_2 \succeq \dots \succeq c_k$,
besides extending $\mathcal{L}_A^{BC}$ (Equation \ref{eq:BCA}) to the ranking loss ${L}_A^{RBC}$ (Equation \ref{eq:rBCA}), we also try to extend $\mathcal{R}_A^{DC}$ to two types of detached constraint ranking loss as follows:

\begin{equation}
    {L}_A^{RDC1} = \log\left[1 + \sum_{c_i \succ c_j } \exp(\mathbf{D}(s_{\theta}^{c_j}) - s_{\theta}^{c_i})\right]
    \label{eq:rRDC1}
\end{equation}

\begin{equation}
    {L}_A^{RDC2} = \log\left[1 + \sum_{c_i \succ c_j, c_j \notin c_{min} } \exp(s_{\theta}^{c_j} - s_{\theta}^{c_i})  +  \sum_{c_i \succ c_{j}, c_j \in c_{min}} \exp(\mathbf{D}(s_{\theta}^{c_{j}}) - s_{\theta}^{c_i})   \right]
    \label{eq:rRDC2}
\end{equation}
where $c_{min}$ is the set of all lowest-quality examples.
Specifically, ${L}_A^{RDC1}$ detachs the score of $c$ when it serves as a negative example, while ${L}_A^{RDC2}$ only detach the score of lowest-quality examples.
We design ${L}_A^{RDC2}$ as we consider that in a ranking scenario, higher-quality examples are inherently constrained by lower-quality ones. Consequently, we hypothesize that constraining only the lowest examples could potentially prevent model degradation.

We also consider a ranking baseline without any constraint:
\begin{equation}
    \mathcal{L}_{A}^R = \log\left[1 + \sum_{c_i \succ c_j} \exp(s_{\theta}^{c_j} - s_{\theta}^{c_i})  \right]
    \label{eq:rBCA}
\end{equation}

\begin{table}[t]
    \centering
    \begin{tabular}{lccccc}
    \toprule
     Methods & $\mathcal{L}_{VFT}$ & +${L}_A^{RBC}$ & +$\mathcal{L}_A^{RDC1}$\ & +$\mathcal{L}_A^{RDC2}$ & +$\mathcal{L}_A^R$\\
     \midrule
     Accuracy & 20.82{$\pm$0.71} & 26.08{$\pm$1.05} & 25.68{$\pm$0.49} & 12.57{$\pm$1.34}  & 7.03{$\pm$0.98}  \\
    \bottomrule
    \end{tabular}
    \caption{Results of LLama-7B on GSM8K fin-tuned by different methods.}
    \label{tab:RDCC}
\end{table}

Table \ref{tab:RDCC} illustrates the results of LLama7B fine-tuned by different methods on GSM8K-RANK.
As is shown:
\textbf{1)}:
The method without setting any constraint $\mathcal{L}_A$ only achieves $7.03$ accuracy, showing the importance of adding a constraint to the alignment loss.
\textbf{2)}:
${L}_A^{RDC2}$, which applies a detached constraint solely to the lowest-quality examples, attains a marginally improved accuracy of $12.57$. However, it also considerably impairs the model's overall performance compared with VFT, indicating that constraining only the lowest-quality examples is insufficient.
\textbf{3)}:
${L}_A^{RDC1}$ is much better than VFT, ${L}_A^{RDC2}$ and $\mathcal{L}_A$, we think the reason is that after detaching all negative scores, 
 ${L}_A^{RDC1}$ prevents the model degradation,
however, it is worse than ${L}_A^{RBC}$,
we hypnosis that ${L}_A^{RDC1}$  only tries to improve all scores, although with different extends, which is not good enough in the ranking situation.

\section{Delve Deeply into  Previous Ranking Losses for Alignment}
\label{sec:analysis_rank}

In this section, we delve deeply into previous widely used ranking losses for alignment, DPO \citep{dpo}, RRHF \citep{yuan2023rrhf} and PRO \citep{song2023preference}, and point out that they all suffer from lack of a constraint term.

Given a ranking sequence $c_1 \succeq c_2 \succeq \dots \succeq c_k$, all ranking losses are proposed to ensure the scores of high-quality examples are larger than those of low-quality examples. 
Ranking losses usually use the token-averaged log-likelihood to represent the score of an example $c$ given by an LLM parameterized by $\theta$:
\begin{equation}
s_{\theta}^{c}=\frac{1}{|c|}{\sum_{j=1}^{|c|} \log P\left(c_j \mid c_{<j}, q;\theta\right)},
\end{equation}

\subsection{DPO}
Direct Preference Optimization (DPO) (the ranking version) optimizes LLMs with the following ranking loss:
\begin{equation}
    \begin{aligned}
    \mathcal{L}_{DPO} &= - \sum_{c_i}\log \frac{ \exp( \beta   s_\theta^{c_i} -  \beta  s_{\theta_{ref}}^{c_i}    ) }{\exp(\beta   s_\theta^{c_i} -  \beta  s_{\theta_{ref}}^{c_i}   ) + \sum_{c_j \prec c_i} \exp(\beta   s_\theta^{c_j} -  \beta  s_{\theta_{ref}}^{c_j}    ) } \\
    & = \sum_{c_i}  \log \left[1 + \sum_{c_j \prec c_i} \exp(   \beta   s_\theta^{c_j} -  \beta  s_{\theta_{ref}}^{c_j} -  \beta   s_\theta^{c_i} +  \beta  s_{\theta_{ref}}^{c_i}    )\right]  
    \end{aligned}
\end{equation}
where $\theta$ and $\theta_{ref}$ are parameters of the training model and reference model, respectively.
The training model and reference model are usually initialized by the same LLM, and DPO freezes the reference model during fine-tuning. $\beta$ is a hyper-parameter of DPO.

To analyze the effectiveness of DPO, we compute the gradient with respect to the parameters $\theta$:

\begin{equation}
\footnotesize
\begin{aligned}
 & \nabla_{\theta} \mathcal{L}_{{DPO}} = -\sum_{c_i} \\
 & \frac{ \sum_{c_j \prec c_i} [\beta\exp(   \beta   s_\theta^{c_j} -  \beta  s_{\theta_{ref}}^{c_j} -  \beta   s_\theta^{c_i} +  \beta  s_{\theta_{ref}}^{c_i}    ) \nabla_{\theta} s_\theta^{c_i} -   \beta\exp(   \beta   s_\theta^{c_j} -  \beta  s_{\theta_{ref}}^{c_j} -  \beta   s_\theta^{c_i} +  \beta  s_{\theta_{ref}}^{c_i}    ) \nabla_{\theta} s_\theta^{c_j} ] }{1 + \sum_{c_j \prec c_i} \exp(   \beta   s_\theta^{c_j} -  \beta  s_{\theta_{ref}}^{c_j} -  \beta   s_\theta^{c_i} +  \beta  s_{\theta_{ref}}^{c_i}    )} 
 \end{aligned}
 \label{eq:dpo}
\end{equation}

Based on $\nabla_{\theta} \mathcal{L}_{{DPO}}$, for each pair $(c_i, c_j)$, $\mathcal{L}_{DPO}$ will decrease the $s_{\theta}^{c_j}$  with  the gradient weight $ \frac{ \beta\exp(   \beta   s_\theta^{c_j} -  \beta  s_{\theta_{ref}}^{c_j} -  \beta   s_\theta^{c_i} +  \beta  s_{\theta_{ref}}^{c_i}    )}{1 + \sum_{c_j \prec c_i} \exp(   \beta   s_\theta^{c_j} -  \beta  s_{\theta_{ref}}^{c_j} -  \beta   s_\theta^{c_i} +  \beta  s_{\theta_{ref}}^{c_i}    )}$, which may lead the model degradation.

In the original DPO paper \citep{dpo}, they observe this catastrophe and alleviate it by setting a very small $\beta$ (e.g., 0.1) to achieve a small gradient weight.
Please refer to the original paper for more details.
However, based on Equation \ref{eq:dpo}, the small $\beta$ also hamper the improvement of positive examples, which may also hinder the model's performance.
Furthermore, solely relying on reducing gradient weights might not be sufficient to prevent model deterioration, as demonstrated in the subsequent analysis of RRHF and PRO.
In this paper, we do not replicate DPO since there is no official public code available for ranking.


\begin{table}[t]
    \centering
    \small
    \begin{tabular}{ccccccccccc}
    \toprule
       Scaling Factor $\beta$ & 0.1 & 0.2 & 0.3 & 0.4 & 0.5 & 0.6 & 0.7 & 0.8 & 0.9 & 1 \\
       \midrule
       Accuracy & 18.75 & 18.01 & 15.05 & 13.20 & 11.79 & 11.79 & 9.83 & 8.78 & 8.62 & 7.51  \\
    \bottomrule
    \end{tabular}
    \caption{The influence of gradient weight scaling factor $\beta$ for RRHF.}
    \label{tab:rrhf_weight}
\end{table}

\subsection{RRHF}
Rank Responses to align Human Feedback (RRHF), which takes candidate ranking into account and distinguishes different candidates through a pair-wise ranking loss:
\begin{equation}
    \mathcal{L}_{RRHF} = \sum_{c_i \succ c_j} \max(0, s_{\theta}^{c_j} - s_{\theta}^{c_i})
\end{equation}

We compute the gradient of $\mathcal{L}_{RRHF}$ with respect to $\theta$:
\begin{equation}
    \nabla_{\theta} \mathcal{L}_{RRHF} = - \sum_{c_i \succ c_j} \left[ \underbrace{\mathbb{I}(s_{\theta}^{c_j} > s_{\theta}^{c_i}) \nabla_{\theta} s_{\theta}^{c_i}}_{\rm increase \  s_{\theta}^{c_i\ }} -  \underbrace{\mathbb{I}(s_{\theta}^{c_j} > s_{\theta}^{c_i}) \nabla_{\theta} s_{\theta}^{c_j}}_{\rm decrease \ s_{\theta}^{c_j}}\right]
\end{equation}

Based on $\nabla_{\theta} \mathcal{L}_{RRHF}$, we can see that although RRHF implicitly introduces a constraint by setting the loss to $0$ when the positive score is larger than the negative score, it still has a drawback:
Whenever $s_{\theta}^{c_j} > s_{\theta}^{c_i}$,  $\mathcal{L}_{RRHF}$ will decrease the $s_{\theta}^{c_j}$  with the same gradient weight $\mathbb{I}(s_{\theta}^{c_j} > s_{\theta}^{c_i})=1$.
This weight might be too large, potentially harming the model's performance.

To illustrate this, we explore the performance of RRHF with a scaling factor $\beta$ on its gradient weight.
As shown in Table \ref{tab:rrhf_weight}, it is evident that as the weight increases (larger $\beta$), the model's performance declines, showing that:
1) The constraint of RRHF is not effective enough to prevent model degradation;
2) We can alleviate the model degradation by making the gradient weight smaller suggested by DPO \citep{dpo};
3) Although we have tried a very small $\beta=0.1$, RRHF still harms the performance, which shows solely relying on reducing gradient weights might not be sufficient to prevent model deterioration.

In fact, in the original RRHF paper \citep{yuan2023rrhf}, the authors have observed that a large ranking weight, such as 10 or 100, significantly impairs model performance, leading them to try a smaller weight (i.e., 1). 
However, they do not analyze the potential reason.
In this paper, we highlight that a key factor causing this discrepancy is the unwarranted reduction of the negative example score, which necessitates imposing a constraint on the ranking loss.
In addition,  we discovered that a weight of 1 can also substantially harm the model's performance in the reasoning task. We believe that the optimal weight of RRHF varies across tasks.


\subsection{PRO}
Preference Ranking Optimization (PRO), which takes candidate ranking into account and distinguishes different candidates through a ranking loss with a dynamic temperature:
\label{sec:pro}
\begin{equation}
    \small
    \mathcal{L}_{PRO} = - \sum_{c_i}  \log \frac{\exp(\tau_i^{max}s_{\theta}^{c_i})}{\exp(\tau_i^{max}s_{\theta}^{c_i})+\sum_{c_j \prec c_i} \exp(\tau_i^js_{\theta}^{c_j})} = \sum_{c_i} \log[1 + \sum_{c_j \prec c_i}\exp(\tau_i^js_{\theta}^{c_j}-\tau_i^{max}s_{\theta}^{c_i}) ]
\end{equation}
\begin{equation}
    \tau_i^j =r^{c_i} - r^{c_j} > 0, \ \ \ \  \tau_i^{max} = \max_{c_j \prec c_i} \tau_i^j
\end{equation}
where $r^{c}$ is the score of $c$ given by a reward model. $\tau_i^j$ is the dynamic temperature for score $s_\theta^{c_j}$.
We compute the gradient with respect to the parameters $\theta$:

\begin{equation}
    \small
  \nabla_{\theta} \mathcal{L}_{PRO} = -\sum_{c_i} \frac{\sum_{c_j \prec c_i} [          \tau_i^{max} \exp(\tau_i^js_{\theta}^{c_j}-\tau_i^{max}s_{\theta}^{c_i}) \nabla_{\theta} s_{\theta}^{c_i}     -        \tau_i^{j} \exp(\tau_i^js_{\theta}^{c_j}-\tau_i^{max}s_{\theta}^{c_i}) \nabla_{\theta} s_{\theta}^{c_j}           ]}{1 + \sum_{c_j \prec c_i}\exp(\tau_i^js_{\theta}^{c_j}-\tau_i^{max}s_{\theta}^{c_i})}
\end{equation}

Based on  $\nabla_{\theta} \mathcal{L}_{PRO}$, we can see that for each pair $(c_i, c_j)$, $\mathcal{L}_{PRO}$ will decrease $s_{\theta}^{c_j}$ with the dynamic gradient weight:
\begin{equation}
\small
{\rm DGW}_i^j = \frac{\tau_i^{j} \exp(\tau_i^js_{\theta}^{c_j}-\tau_i^{max}s_{\theta}^{c_i})}{1 + \sum_{c_j \prec c_i}\exp(\tau_i^js_{\theta}^{c_j}-\tau_i^{max}s_{\theta}^{c_i})},
\end{equation}
which may harm the model's performance.
However, the dynamic gradient weight that is computed based on the reward is more reasonable than the constant value of $1$ used in RRHF, and thus PRO outperforms RRHF.
Specifically, when there is a substantial reward gap between higher-quality and lower-quality, indicated by a large value $\tau_i^j$. 
It is reasonable to increase the penalty for negative example scores (large ${\rm DGW_i^j}$), and vice versa.
To demonstrate this, we remove the dynamic temperature term, i.e., $\tau_j^j$ and $\tau_i^{max}$, from PRO.
As shown in Table \ref{tab:pro}, we can see that PRO significantly outperforms PRO (remove $\tau$) when there is no constraint.
However, the performance gap shrinks when adding our detached constraint.
These results indicate:
1) To a certain extent, the dynamic temperature's effectiveness stems from its ability to make PRO reduce the negative score in a more reasonable manner.
2) The dynamic temperature is useful to prevent model degradation but is not good enough.

\begin{table}[t]
    \centering
    \begin{tabular}{lcc|cc}
    \toprule
        Methods & PRO & PRO (remove  $\tau$) & PRO + RDC1 & PRO (remove  $\tau$) + RDC1 \\
        \midrule
        Accuracy & 18.73{$\pm$0.31} & 7.18{$\pm$0.78} & 25.84{$\pm$0.48} & 25.43{$\pm$0.98}  \\
    \bottomrule
    \end{tabular}
    \caption{The importance of dynamic temperature of PRO. ``remove $\tau$'' denotes remove the dynamic temperature term, i.e., $\tau_j^j$ and $\tau_i^{max}$ from PRO. ``+RDC1'' denotes add our ranking detach technical (Equation \ref{eq:rRDC1}).}
    \label{tab:pro}
\end{table}

\subsection{Summary}
Our analysis reveals that previous ranking-based alignment works have some limitations:

1) Although their methods consist of some strategies to prevent model degradation (i.e., using a scaling factor to reduce gradient weight for DPO, setting the loss to 0 for RRHF, and incorporating a dynamic temperature for PRO), they do not recognize the importance of constraints for ranking-based alignment methods in their papers. 

2) Their strategies essentially involve diminishing the gradient weight's magnitude, which is proven to be insufficient (at least in the reasoning tasks). 

Different from previous works, in this paper:

1) We point out an assessment misalignment problem of VFT for reasoning and highlight the importance of constraint for alignment to prevent model degradation.

2) We introduce a  novel constraint alignment loss. The constraint alignment loss with a boundary constraint term not only alters the magnitude but also adjusts the direction of the gradient weight depending on the condition, which is proven to be very effective in preventing model degradation and enhancing the reasoning ability of LLMs.

\section{Case Study}
\label{sec:case}
\begin{table*}[!tp]
\setlength\tabcolsep{4pt}

\centering
    \begin{tabular} {p{2.2cm}|p{10.7cm}}
        \toprule
         \textbf{Question} & \textit{At the end of the school year, Kate asked her teacher if she could have the 3 boxes of 64 crayons since they were all worn down to small pieces.  The teacher agreed.  Kate took the crayons home and decided to make crayon muffins by melting 8 small pieces of crayons together in a muffin tin that she placed in an oven for a few minutes. If she can sell her muffin crayons for \$1.50 each, how much money can Kate make?}
         \\
         \midrule
         \textbf{Answer} & She was able to get 3 boxes of 64 crayons so she has \textcolor{blue}{3*64 = $<<$3*64=192$>>$192} crayons
Each crayon muffin needs 8 crayons and she has 192 crayons so she can make \textcolor{blue}{192/8 = $<<$192/8=24$>>$24} crayon muffins
If she makes 24 crayon muffins and sells them for \$1.50 apiece, she can make \textcolor{blue}{24*1.5 = \$$<<$24*1.5=36.00$>>$36.00}
[ANS] 36.  \\
         \midrule
         \midrule
         \textbf{VFT} & She has 3 boxes of crayons that are all worn down to small pieces so she has 3*64 = \textcolor{blue}{$<<$3*64=192$>>$192} pieces of crayons
She melts 8 pieces of crayons together in a muffin tin so she has \textcolor{red}{8+192 = $<<$8+192=200$>>$200} crayon muffins
She can sell her muffin crayons for \$1.50 each so she can make \textcolor{red}{200*1.5 = \$$<<$200*1.5=300$>>$300}
[ANS] 300.
         \\
        \midrule
        \textbf{AFT} &Kate has 3 boxes * 64 crayons = \textcolor{blue}{$<<$3*64=192$>>$192} crayons.
She melts 192 crayons / 8 crayons per muffin = \textcolor{blue}{$<<$192/8=24$>>$24} muffins.
She sells the muffins for \$1.50 each and has 24 muffins, so she makes 24 muffins * \$1.50/muffin = \textcolor{blue}{\$$<<$24*1.5=36$>>$36.}
[ANS] 36.
         \\
         \midrule
         \textbf{AFT (\textit{remove} constraint)} & She can make 8 pieces of crayons per muffin tin so 3 boxes * 8 pieces/tin = 8 * 8 = $<<$3*8=24$>>$24 pieces fit in 1 tin
Each melted crayon is worth \$1.50 because 24 pieces * 1.50/64 pieces = \textcolor{red}{1.50/1.50/24 pieces = 1.50/1.50}
She can make 4 tins of melted crayons because 24 pieces \textcolor{red}{/ 1.50 / 1.5 / 24 pieces / 1.50 / 1.50 / 24 pieces / 1.50 / 1.50 / 24 pieces / 1.50 / 1.50 / 24 pieces / 1.50 / 1.50 / 24 pieces / 1.50 / 1.50 / 24 pieces / 1.50 / 1.50 / 24 pieces / 1.50 / 1.50} \\
         \bottomrule
    \end{tabular}
    \caption{A case study to intuitively show the effectiveness of AFT with boundary constraint.  the right and wrong steps are colored by  \textcolor{blue}{blue} and \textcolor{red}{red}, respectively.}
    \label{tab:case}
\end{table*}

We also conducted a case study to intuitively show the importance of our constraint alignment.
As shown in Table \ref{tab:case}, given the question, our AFT successfully gives the correct COT and answer, while VFT gives the wrong COT at the second step (colored red), demonstrating the superiority of AFT.
More importantly, after removing the boundary constraint, the generative ability of LLM seems to degrade, resulting in outputting many repeat and meaningless output tokens.

\end{document}